%% file: kvae.tex
\documentclass{article}

\PassOptionsToPackage{numbers, compress}{natbib}

\usepackage[final]{neurips_2024}
\usepackage[utf8]{inputenc} 
\usepackage[T1]{fontenc}    
\usepackage{url}            
\usepackage{booktabs}       
\usepackage{amsfonts}       
\usepackage{nicefrac}       
\usepackage{microtype}      
\usepackage{xcolor}         
\usepackage{url}
\usepackage{graphicx}
\usepackage{svg}
\usepackage{animate}
\usepackage{listings}
\usepackage{algorithm}
\usepackage{algpseudocode}
\usepackage{enumitem}
\usepackage{silence}
\usepackage{makecell}
\usepackage{graphicx}  
\usepackage{wrapfig}   
\usepackage{caption}
\usepackage{subcaption}
\usepackage{float}
\usepackage{multirow}
\usepackage[accsupp]{axessibility}  

\usepackage[pagebackref,breaklinks,colorlinks,citecolor=gray]{hyperref}

\makeatletter
\renewcommand{\@noticestring}{\@neuripsordinal.}
\makeatother

\title{KVAE-3.0: Extending Tokenizers Capabilities with Latent Alignment}

\author{%
  Ivan Kirillov\textsuperscript{3}\\
  Kandinsky Lab\\
  \And
  Denis Parkhomenko\textsuperscript{1, 2, 3}\\
  Kandinsky Lab\\
  \And
  Andrey Shutkin\textsuperscript{1}\\
  Kandinsky Lab\\
  \And
  Kirill Chernyshev\textsuperscript{1}\\
  Kandinsky Lab\\
  \And
  David Chikovani\textsuperscript{2}\\
  Kandinsky Lab\\
  \And
  Kirill Malakhov\textsuperscript{2}\\
  Kandinsky Lab\\
  \And
  Azat Saginbaev\textsuperscript{3}\\
  Kandinsky Lab\\
  \And
  Alexander Ivanov\textsuperscript{3}\\
  Kandinsky Lab\\
  \And
  Ivan Mikheev\textsuperscript{1}\\
  Kandinsky Lab\\
  \And
  Egor Silvestrov\textsuperscript{3}\\
  Kandinsky Lab\\
  \And
  Valeria Kobenko\textsuperscript{2}\\
  Kandinsky Lab\\
  \And
  Ilya Vasiliev\textsuperscript{2}\\
  Kandinsky Lab\\
  \And
  Ilya Trushkin\textsuperscript{2}\\
  Kandinsky Lab\\
  \And
  Vladimir Chernyy\textsuperscript{3}\\
  Kandinsky Lab\\
  \And
  Konstantin Zakharov\textsuperscript{2}\\
  Kandinsky Lab\\
}

\newif \ifhq
\hqfalse

\begin{document}

\maketitle
\footnotetext[1]{worked on video tokenization}
\footnotetext[2]{worked on image tokenization}
\footnotetext[3]{worked on audio tokenization}

\begin{abstract}
State-of-the-art results in multimodal generation are achieved by Latent Diffusion Models (LDM), which rely on continuous representations of source media.
Such representations should carry features of original data distribution, be compact enough for later computations and overall denoising predisposition (e.g. in flow matching approach).
This report presents KVAE-3.0, set of tokenizers for audio/image/video domains with latent spaces possessing such properties, what is shown in series of experiments.
Specifically, KVAE-Audio-Stereo is a 48 kHz VAE for general stereo audio at a 50 Hz latent rate, exceeding 70\% win-rate on generations quality comparison against leading opensource VAEs: Stable Audio 3 and LTX-2.5.
An image model KVAE-2D-3.0 matches latent size of FLUX.2-VAE, obtains higher PSNR and SSIM on reconstruction, but also provide faster FID-convergence and even outperform the latter in side-by-side evaluation.
KVAE-3D-3.0 compress frames spatially by factor of 32, temporally by 4 and outperforms MiniMax-H3 VAE over human preference test on generations by all criteria.
Inference code and link to model weights are publically available at \url{https://github.com/kandinskylab/kvae}.
\end{abstract}


\section{Introduction}
\label{sec:intro}
\input{content/introduction}

\section{Related Works}
\label{sec:related_works}
\input{content/related_works}

\section{Visual tokenizers}
\label{sec:visual_tokenizers}

\input{content/visual/overview}
\input{content/visual/reconstruction}
\input{content/visual/generation}

\section{Audio tokenizer}
\label{sec:audio_tokenizer}

\input{content/audio/overview}

\input{content/audio/reconstruction}
\input{content/audio/generation}

\section{Conclusion}
Models, presented in the report, have different domains, yet during their development we've found universal alignment techniques, which boost their generation capabilites. Evaluations in corresponding sections confirm that statement and allow to place tokenizers among best opensource models. Moreover, we opensourced weights and code under the MIT license. 

\vspace{30pt}
{
\bibliographystyle{plain}
\bibliography{kvae,content/audio_refs}
}

\end{document}

%% file: content/introduction.tex
Latent diffusion has established a common computational framework for the generation of otherwise very different signals. By separating signal compression from generative modeling, latent diffusion and related flow-matching models have made it possible to synthesize high-resolution images, extended video sequences and long-form audio without modeling their raw samples directly~\citep{rombach2022ldm,blattmann2023videoldm,liu2023audioldm,evans2026stableaudio3}. Recent systems have moved beyond generation within individual modalities: Veo 3.1 and LTX-2.5 synthesize video with synchronized audio from textual prompts and visual references~\citep{deepmind2025veo,lightricks2026ltx25}. Notably, joint generation does not require a single tokenizer; LTX-2, for example, operates on separate audio and video representations~\citep{hacohen2026ltx2}.

These representations are typically learned by a first-stage autoencoder, often a variational autoencoder (VAE), which we refer to as a tokenizer. Besides determining the sequence length and the information available to the decoder, the tokenizer defines how the underlying structure of the data is exposed to the generative model. Conventional reconstruction objectives provide only indirect supervision for this latter property: evaluating each candidate representation through downstream diffusion training would be prohibitively expensive. Foundation models offer an alternative source of supervision. Their representations capture relationships between image regions, video events or acoustic patterns that need not be explicitly organized by a reconstruction objective. Aligning the tokenizer's latents with such representations has been shown to facilitate subsequent generative modeling~\citep{yao2025vavae,niu2025semanticvae,zhan2026omnivae}. This suggests that the benefits of representation learning can be transferred to generative tokenizers without requiring a common representation space across modalities.

In report we share details regarding such tokenizanation models for audio, image and video domain, all of which explored latent alignment as technique for better subsequent generative modeling. We show via comparison by objective metrics and human preference test superiority of this models to leading opensource solutions. This work should be considered as continuation of ~\citep{kvaereport}, since backbone (in training, architecture, data) is shared, but there is also signifant updates, expanding their functionality: stereo audio processing (KVAE-Audio-Stereo), increased compression ratio for video model (KVAE-3D-3.0) and improved latent space for image tokenizer (KVAE-2D-3.0), enabling faster training of generative DiT ~\citep{kandinsky5}. Content organized as follows: relevant to this report works outlined in Sec. ~\ref{sec:related_works}. Visual models (image and video) are introduced in Sec. ~\ref{sec:visual_tokenizers} with their results on reconstruction and generation tasks. Last part, Sec. ~\ref{sec:audio_tokenizer} is dedicated to KVAE-Audio-Stereo.

%% file: content/related_works.tex
\subsection{Visual tokenizers for latent diffusion models}

Image tokenizers have evolved towards higher information capacity at a fixed spatial downsampling factor. The autoencoders of FLUX.1 and FLUX.2~\citep{bfl2024flux,bfl2025flux2} increase the latent width of the original Stable Diffusion tokenizer from 4 channels to 16 and 32, respectively. Importantly, the latter combines increased capacity with semantic regularization and evaluates its latent space in the context of both reconstruction and generative modeling. For video, where the sequence length also depends on temporal resolution, recent developments emphasize spatiotemporal compression. The causal VAEs of HunyuanVideo-1.5 and the 5B variant of Wan2.2~\citep{wu2025hunyuan15,wan2025wan22} compress the spatial axes by a factor of 16 and the temporal axis by a factor of 4. Beyond increasing compression, the design of video tokenizers therefore involves retaining the temporal structure of the signal in a substantially reduced sequence of latent tokens. MiniMax-H3~\citep{minimax2026h3} adopts the same spatiotemporal compression factors with 24 latent channels and explicitly optimizes its visual latent space for both reconstruction and generative modeling. The applicability of video tokenizers extends beyond media synthesis: FLUX 3 Action~\citep{bfl2026flux3action} employs a frozen video VAE in a world-action model that jointly predicts future video frames and robot actions.

\subsection{Audio tokenizers for latent diffusion models}

Continuous audio tokenizers differ in the signal representation they compress and the temporal resolution at which they expose it to the generator. Whereas AudioLDM~\citep{liu2023audioldm} models latents of mel-spectrograms, Stable Audio Open~\citep{evans2024sao} directly encodes 44.1 kHz stereo waveforms at a latent rate of approximately 21.5 Hz. More recent stereo waveform tokenizers include the VAE of ACE-Step 1.5~\citep{gong2026acestep,gong2026acevae}, operating at 48 kHz with a latent rate of 25 Hz, and SAME~\citep{parker2026same}, the autoencoder of Stable Audio 3~\citep{evans2026stableaudio3}, which uses transformer-based resampling to reduce the latent rate to approximately 10.8 Hz at 44.1 kHz. In contrast, the stereo audio VAE of LTX-2.5~\citep{lightricks2026ltx25,lightricks2026ltxcore} retains a mel-spectrogram interface and relies on a separate vocoder to produce the waveform. These designs illustrate different allocations of signal modeling between the tokenizer, the latent generator and the waveform synthesis stage.

\subsection{Alignment with external representations}

The use of pretrained representations in diffusion modeling has been explored at different stages of the generative pipeline. REPA~\citep{yu2025repa} supervises intermediate denoiser features using representations of clean images, whereas VA-VAE~\citep{yao2025vavae} applies supervision directly to the tokenizer latents, matching both feature directions and pairwise token relationships. MAETok~\citep{chen2025maetok} obtains semantically structured latents through masked prediction of multiple feature targets, and REPA-E~\citep{leng2025repae} uses denoiser-side alignment to jointly optimize the tokenizer and diffusion transformer. A complementary approach is to build the tokenizer around the representation model itself: RAE and VFM-VAE~\citep{zheng2026rae,bi2026vfmvae} retain a frozen foundation encoder, while AlignTok~\citep{chen2026aligntok} adapts it under an additional semantic preservation objective.

Extensions to temporally structured signals make the choice of representation and alignment objective particularly relevant. DeRA~\citep{guo2025dera} separately aligns appearance and motion tokens with image and video foundation models for discrete, autoregressive video generation, while VideoRAE~\citep{xie2026videorae} compresses features of a frozen video encoder into latents suitable for diffusion and autoregression. In speech synthesis, Semantic-VAE~\citep{niu2025semanticvae} aligns continuous latents with pretrained speech features, and JMAS-VAE~\citep{cheng2026jmas} investigates alignment along both temporal and feature axes, as well as their joint formulation. For broader audio content, SAME~\citep{parker2026same} combines acoustic feature regression, contrastive supervision and an auxiliary diffusion objective. More recently, OmniVAE~\citep{zhan2026omnivae} combines modality-specific feature distillation with audio-video contrastive alignment for joint generation.

%% file: content/visual/overview.tex
\subsection{Overview}
\begin{figure}[H]
\centering
\includegraphics[width=\linewidth]{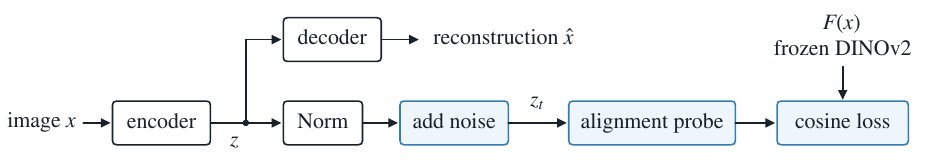}
\caption{Joint training of KVAE-2D-3.0. The alignment probe is discarded after training.}
\label{fig:image-training-scheme}
\end{figure}

KVAE-2D-3.0 is an \emph{f8} variational autoencoder with 32 latent channels.
Our alignment probe follows the architecture of the first eight blocks of the SiT-XL/2 diffusion transformer used to train KVAE-2D-2.1 with REPA-E.
We train the probe from scratch for feature alignment and omit the diffusion model's velocity-prediction loss.
Given a noise-corrupted latent sample and its noise level, the probe predicts patch features extracted from the clean image by a frozen DINOv2 ViT-B/14 encoder~\citep{dinov2}.
The alignment loss updates both the probe and the VAE encoder (Figure~\ref{fig:image-training-scheme}).

At the same time, video tokenizer KVAE-3D-3.0 is an \emph{f32t4} variational autoencoder with 128 latent channels, so its compression ratio totally is 2 times more than predecessor, KVAE-3D-2.0 with \emph{f16t4}. Main difference between these two models is in training strategy: new model adds alignment on V-JEPA ~\citep{vjepa} features, which impoves semantic capacity of latents.

\paragraph{Representation alignment in KVAE-2D-2.1}
We trained KVAE-2D-2.1~\citep{kvae21,kvaereport} and a 28-block SiT-XL/2 diffusion transformer~\citep{sit} from scratch with REPA-E~\citep{repae}. Each update used 256 crops of $256\times256$ pixels sampled from our internal dataset of 500 million images, without text or class conditioning. After 400{,}000 joint training steps, we froze the encoder and fine-tuned the decoder on $512\times512$ crops.

Following REPA~\citep{repa}, we placed an alignment head after block~8 of the transformer to predict a DINOv2 feature for each patch. During VAE updates, the transformer weights were fixed and the alignment loss, weighted by 1.5, supplied a gradient to the encoder through those eight blocks. During transformer updates, we detached the latents and optimized velocity prediction along the linear path from latent to noise, together with the alignment loss weighted by 0.5. This separation lets us retain just the eight blocks needed for alignment in KVAE-2D-3.0. We train the VAE and probe from scratch and discard the probe after training.

\paragraph{Tokenizer architecture}
KVAE-2D-3.0 retains the spatial compression and latent channel count of its predecessor. We reduce the width of the deepest residual stage from 1{,}024 to 512 channels. The four stages have widths of 128, 256, 512, and 512 channels. While KVAE-2D-2.1 was purely convolutional, KVAE-2D-3.0 uses self-attention in both the encoder and decoder bottlenecks. The tokenizer used at inference comprises a 34.8M-parameter encoder and a 50.8M-parameter decoder, for a total of 85.6M parameters compared with 218.4M in KVAE-2D-2.1. The two versions differ by less than 0.04\,dB in reconstruction PSNR on COCO val2017 (Table~\ref{tab:img_reconstruction}).

With increase in spatial compression ratio of video tokenizer (\emph{f16t4} of KVAE-3D-2.0 vs. \emph{f32t4} of KVAE-2D-2.0) new blocks were added to encoder and decoder: one downsample, one upsample and few residual blocks. Even with such high compression, whole model stays convolutional, so no spatial tiling is necessary on inference even for FullHD resolution. 

\paragraph{Training image encoder with noisy latents}
The decoder reconstructs image $x$ from a sample $z$ of the diagonal Gaussian encoder posterior $q_\phi(z\mid x)$. On the alignment branch, we apply channelwise batch normalization to the same sample and interpolate the normalized latent $\widetilde z$ with Gaussian noise to obtain
\begin{equation}
 z_t=(1-t)\,\widetilde z+t\,\epsilon,\qquad
 t\sim\mathcal U(0,1),\qquad \epsilon\sim\mathcal N(0,I).
 \label{eq:image-corruption}
\end{equation}
The probe embeds $2\times2$ patches of the $32\times32$ latent grid. For DINOv2, we resize the same crop to $224\times224$ and extract features from $14\times14$ image patches. Both operations yield a $16\times16$ grid, so each probe prediction has a teacher target at the corresponding spatial position. Each probe block has width 1{,}152, uses 16 attention heads, and is conditioned on $t$. A multilayer perceptron projects its final output at each position to the 768-dimensional teacher feature space.

Let $p_i(z_t,t)$ be the projected probe feature at position $i$ and $d_i(x)$ the frozen teacher feature there. With $N=256$ patch positions, we minimize
\begin{equation}
 \mathcal L_{\mathrm{align}}
  =-\frac{1}{N}\sum_{i=1}^{N}
       \frac{p_i(z_t,t)^\top d_i(x)}
            {\lVert p_i(z_t,t)\rVert_2\,\lVert d_i(x)\rVert_2},
 \qquad
 \mathcal L=\mathcal L_{\mathrm{VAE}}+1.5\,\mathcal L_{\mathrm{align}} .
 \label{eq:image-align}
\end{equation}
The VAE objective $\mathcal L_{\mathrm{VAE}}$ combines pixelwise $L_1$ and LPIPS~\citep{lpips} with unit weights, a KL divergence to a standard Gaussian prior weighted by $10^{-6}$, and an adversarial term weighted by~0.1. Overall reconstruction training strategy based on previous works ~\citep{dcaenvidia, vivat}.

We developed our alignment probe independently of RoAR-VAE~\citep{yue2026roar}, whose RobustAlign objective also predicts semantic features of the clean image from noisy latents.

We jointly train the tokenizer and probe for 600{,}000 steps on $256\times256$ crops from our internal dataset.

\paragraph{Decoder fine-tuning}
For KVAE-2D-3.0 we then freeze the encoder and fine-tune the decoder on random $512\times512$ crops. For a crop $x$ and its reconstruction $\hat x$, the decoder objective is
\begin{equation}
 \mathcal L_{\mathrm{dec}}
 = \mathrm{MAE}(x,\hat x)
   + 0.2\,\mathrm{LPIPS}_{\mathrm{VGG}}(x,\hat x)
   + 5\times10^{-4}\,\mathcal L_{\mathrm{adv}}(\hat x).
 \label{eq:image-decoder}
\end{equation}
The adversarial term uses a four-layer patch discriminator.

We use Adam with learning rates of $5\times10^{-5}$ for the decoder and $10^{-4}$ for the discriminator, a global batch of 24 crops, and bfloat16 mixed precision. After 5{,}000 discriminator warm-up steps, we alternate decoder and discriminator updates, for a total of 7{,}500 decoder updates and 12{,}500 discriminator updates.

Video model, KVAE-3D-3.0, finetunes its decoder not only on images, but on video also. With slightly different loss, main finding is the same: best results obtained with higher resolution. 

%% file: content/visual/reconstruction.tex
\subsection{Reconstruction results}

We evaluate (see Table~\ref{tab:img_reconstruction}) on all 5,000 COCO val2017 images ~\citep{cocodataset} at native scale using the KVAE-2D-2.1 protocol ~\citep{kvaereport}. We center-crop each dimension to a multiple of 32 and decode the posterior mean in bfloat16. We average peak signal-to-noise ratio (PSNR), structural similarity (SSIM), and AlexNet LPIPS over images.

\begin{table}[H]
\centering
\caption{Image reconstruction on COCO val2017.
Arrows indicate the preferred direction.}
\label{tab:img_reconstruction}
\begin{tabular}{lccccc}
\toprule
Tokenizer & Latent channels & Total parameters & PSNR (dB) $\uparrow$ & SSIM $\uparrow$ & LPIPS $\downarrow$ \\
\midrule
FLUX.1-dev   & 16 & 83.8 M & 29.89 & 0.8662 & 0.0234 \\
FLUX.2-dev   & 32 & 84.0 M & 31.37 & 0.8924 & \textbf{0.0167} \\
KVAE-2D-2.1  & 32 & 218.4 M & 32.15 & 0.9057 & 0.0264 \\
KVAE-2D-3.0  & 32 & 85.6 M & \textbf{32.18} & \textbf{0.9070} & 0.0258 \\
\bottomrule
\end{tabular}
\end{table}

Video models were evaluated on original videos from MCL-JCV ~\citep{mcljcv} of 720p and cropped in accordance with spatial, temporal factors. Since every computed metric is frame-wise we average, but in two steps: first for across all frames in video, second across obtained numbers. Different from image tokenizers comparison, here we use LPIPS over VGG.

\begin{table}[H]
\centering
\caption{Video reconstruction on MCL-JCV 720p}
\label{tab:vid_reconstruction}
\begin{tabular}{lccccc}
\toprule
Tokenizer & Compression factor & Channels & PSNR (dB) $\uparrow$ & SSIM $\uparrow$ & LPIPS $\downarrow$ \\
\midrule
FLUX.3-Action   & \emph{f32t4} & 96 & \textbf{32.79} & \textbf{0.878} & 0.181 \\
MiniMax-H3   & \emph{f16t4} & 24 & 32.44 & 0.877 & 0.176 \\
KVAE-3D-3.0  & \emph{f32t4} & 128 & 31.72 & 0.866 & \textbf{0.175} \\
\bottomrule
\end{tabular}
\end{table}

%% file: content/visual/generation.tex
\subsection{Generation results}

Since both models intended for further use by LDM, reconstruction metrics are not enough. Therefore, experiments with generation models were conducted to compare latent spaces among different VAEs. Conditions are same for all runs and the only variable left is VAE itself. Video generation models follow image training as its first stage, which is possible due to causality of VAEs, with video training on later steps.\par

\begin{figure}[H]
\centering
\includegraphics[width=.83\linewidth]{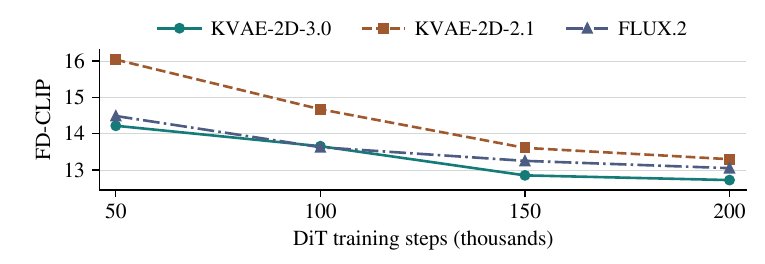}
\caption{Generation FD-CLIP at the four DiT training steps evaluated for all three tokenizers. Lower values indicate closer agreement with the reference distribution. Evaluation used 3{,}000 English prompts, $256\times256$ images, 50 sampling steps, guidance scale 5, and sampling shift 5.}
\label{fig:dit-fd-clip}
\end{figure}

\textbf{Generation quality during DiT training.} We trained separate Kandinsky~5 text-to-image DiTs~\citep{kandinsky5} with 2B parameters on latents from KVAE-2D-3.0, KVAE-2D-2.1, and FLUX.2. All three runs used 64 GPUs. We evaluated generation quality by the Fr\'echet distance between generated and reference images in CLIP ViT-L/14 feature space (FD-CLIP)~\citep{clip}. At 100K steps, FLUX.2 and KVAE-2D-3.0 had similar scores of 13.63 and 13.66, respectively (Figure~\ref{fig:dit-fd-clip}). At 200K steps, KVAE-2D-3.0 had the lowest FD-CLIP of the three (12.72), followed by FLUX.2 (13.05) and KVAE-2D-2.1 (13.30).\par

\begin{figure}[H]
\centering
\includegraphics[width=.74\linewidth]{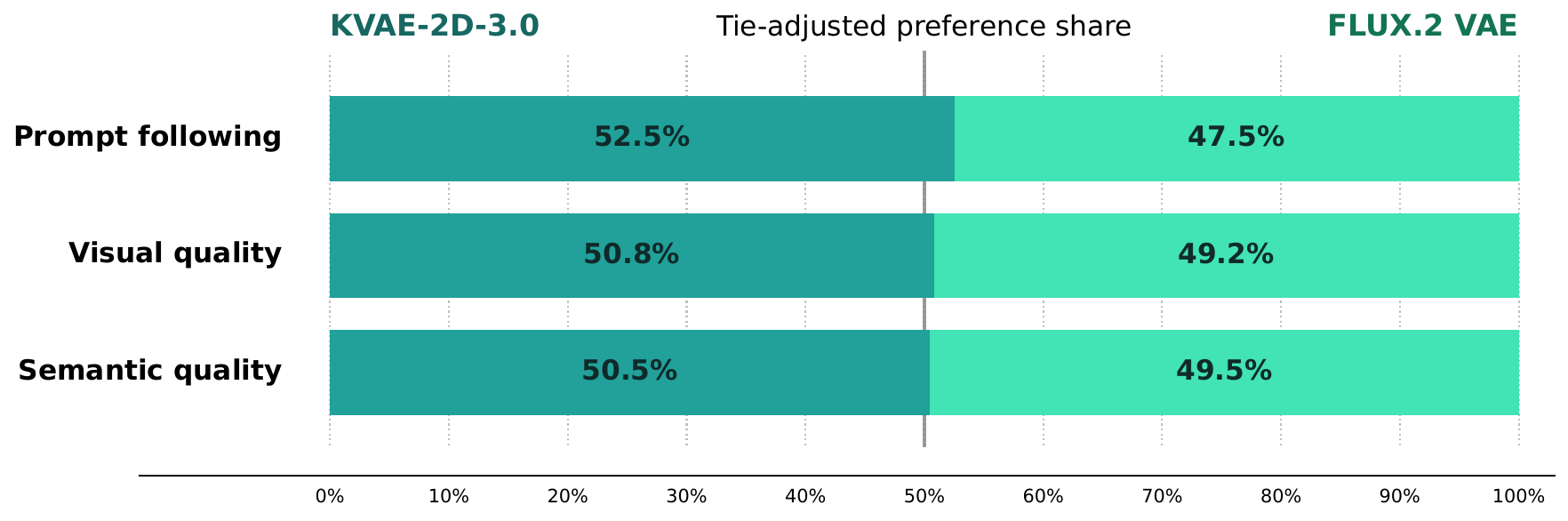}
\caption{Tie-adjusted preference shares by criterion. We split credit equally for \emph{both good} ratings and exclude \emph{both poor} ratings when calculating the shares.}
\label{fig:human-preferences}
\end{figure}

\textbf{Human preferences.} We compared the KVAE-2D-3.0 and FLUX.2 generators at 200K DiT steps. We used the same prompt and random seed for both images in each pair. The KVAE-based generator's tie-adjusted preference share was highest for prompt following (52.5\%) and close to 50\% for visual and semantic quality (Figure~\ref{fig:human-preferences}).\par

In the same setup were compared KVAE-3D-3.0 and MiniMax-H3 VAE with only difference in patch size. MiniMax-H3 use patch of 2x2, while KVAE-3D, having higher spatial compression, utilizes only size 1x1. That being said, dimensionality of input patch vector of KVAE-3D is higher: 128 channels against 96 of MiniMax-H3.

\begin{figure}[H]
\centering
\includegraphics[width=.74\linewidth]{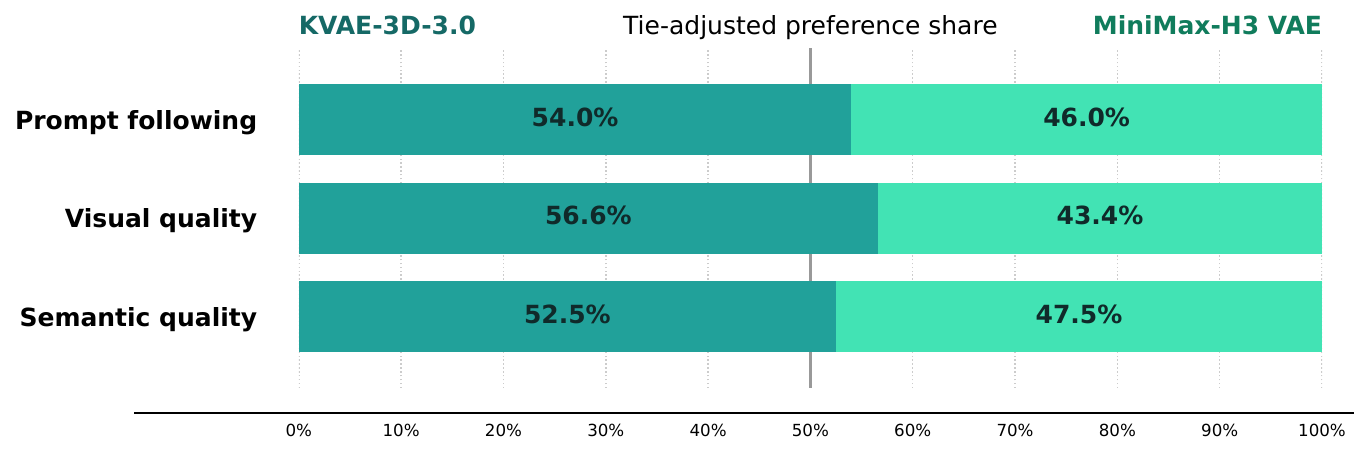}
\caption{Human preferences for video models}
\label{fig:human-preferences-vid}
\end{figure}

\textbf{DiffusionBench.} On the text-to-image track (Table~\ref{tab:generation}), the generator trained on KVAE-2D-3.0 latents achieves the highest GenAI-Bench score (0.726) and the second-highest three-metric mean (0.745), behind E2E-Qwen-Image-VAE (0.747).\par

\begin{table}[H]
\centering
\caption{DiffusionBench text-to-image scores. We evaluate the 0.61B-parameter KVAE-based generator using exponential moving average weights at 100K steps. Baselines follow~\citep{diffusionbench2026}. We show three of the 12 tokenizer configurations used for ranking. The ranking excludes the separate FLUX.2-VAE + REG configuration.}
\label{tab:generation}
\begin{tabular}{lcccc}
\toprule
Tokenizer & GenEval & DPG-Bench & GenAI-Bench & Mean \\
\midrule
E2E-Qwen-Image-VAE & \textbf{0.691} & 0.835 & 0.714 & \textbf{0.747} \\
KVAE-2D-3.0        & 0.673 & \textbf{0.837} & \textbf{0.726} & 0.745 \\
FLUX.2-VAE         & 0.675 & 0.830 & 0.712 & 0.739 \\
\bottomrule
\end{tabular}
\end{table}

%% file: content/audio/overview.tex
\subsection{Overview}

Our model is a waveform VAE based on the Descript Audio Codec architecture, designed for stereo signals.
The encoder takes two waveform channels at 48 kHz and downsamples with five residual blocks of strides (2, 3, 4, 5, 8), a total hop of 960 samples. Left and right are encoded jointly: the latent is not a concatenation of two mono streams, and each frame carries inter-channel information in its feature dimensions. While developing, we found that simply changing input domain from mono to stereo, leaving  baseline architecture ~\citep{kvaereport} with 64 latent channels, raises MEL from 0.76/0.74/0.55 to 0.93/0.89/1.66 for sounds, music, and speech, and lowers SI-SDR by 3.6 and 5.3 dB on sounds and music (see Table~\ref{tab:stereo_reconstruction}). But extending channel dimension to 128 resulted in degradation of generations, although stabilized reconstruction.

\begin{table}[H]
\centering
\caption{Reconstruction on the stereo test set for the three
configurations of the problem statement.
The models differ only in the latent width and in mono
versus stereo input and output.}
\label{tab:stereo_reconstruction}
\begin{tabular}{llrrr}
\toprule
Metric & Model & Sounds & Music & Speech \\
\midrule
\multirow{3}{*}{MEL$\downarrow$}
& Mono-64    & 0.7591 & 0.7413 & 0.5481 \\
& Stereo-64  & 0.9291 & 0.8862 & 1.6555 \\
& Stereo-128 & 0.8277 & 0.7936 & 0.441  \\
\midrule
\multirow{3}{*}{STFT$\downarrow$}
& Mono-64    & 2.7083 & 2.9303 & 1.6684 \\
& Stereo-64  & 2.8247 & 2.9894 & 4.0507 \\
& Stereo-128 & 2.8103 & 2.9818 & 1.400  \\
\midrule
\multirow{3}{*}{SI-SDR$\uparrow$}
& Mono-64    & 1.168  & 11.250 & 9.077  \\
& Stereo-64  & -2.411 & 5.974  & 9.339  \\
& Stereo-128 & -5.996 & 8.606  & 10.679 \\
\bottomrule
\end{tabular}
\end{table}

Due to the aforementioned challenges, we've modified alignment with foundational model and add stereo-specific losses. Explicitly, alignment done to a subspace of teacher features instead of learning a projection and use the audio encoder of a perceptual-quality predictor rather than a speech-oriented self-supervised encoder. Teacher audio is resampled to 16 kHz and processed independently for the left and right channels. As for spectral reconstruction, it is evaluated in two channel bases separately: left/right ($M = (L + R)/2$) and mid/side ($S = (L-R)/2$). We also add the phase objective of SAME, $\mathcal{L}_{\mathrm{IFGD}}$, and an additional joint-stereo adversarial objective with discriminator perceiving both channels.

%% file: content/audio/reconstruction.tex
\subsection{Reconstruction results}

Reconstruction metrics are reported on a test set of music, speech, and sound effects assembled from open datasets, filtered by stereo status and true sample rate, and disjoint from the training data (about 13 hours in total). We report multi-scale mel (MEL) and STFT distances, MFCC distance, waveform $L^1$, SI-SDR, SNR, and, for speech, PESQ. Additionaly we add cross-channel phase coherence (CCPC), the energy-weighted phasor coherence between the inter-channel phase differences of reference and reconstruction, averaged over four STFT resolutions. It measures the stereo image directly, which per-channel metrics do not. Characteristics of all tested models provided in Table~\ref{tab:audio_vae_interfaces}.

\begin{table}[H]
\centering
\caption{Interfaces of the chosen audio VAEs.
Latent rate is frames per second of audio.}
\label{tab:audio_vae_interfaces}
\begin{tabular}{lrrrrrl}
\toprule
Model & Sample rate & Channels & Latent rate
& $d_z$ & Params & Representation \\
\midrule
SAME-L
& 44.1\,kHz & 2 & $\sim$10.8\,Hz & 256 & 852M & waveform \\
LTX 2.5 audio VAE
& 48\,kHz & 2 & 25\,Hz & 128 & 182M & mel + vocoder \\
Ours
& 48\,kHz & 2 & 50\,Hz & 128 & 167M & waveform \\
\bottomrule
\end{tabular}
\end{table}

Table~\ref{tab:stereo_reconstruction_44} reports results at 44.1\,kHz, and Table~\ref{tab:stereo_reconstruction_48} reports results at 48\,kHz. Results are grouped by the sample rate at which the metrics are computed. Within each group, the reference and all reconstructions are downsampled to that rate. LTX 2.5 analyzes a 16\,kHz mel representation and is evaluated on its 48\,kHz vocoder output, so its metrics characterize the complete analysis--VAE--vocoder pipeline rather than direct full-band transmission through the latent.

\begin{table}[H]
\centering
\caption{Stereo reconstruction results at 44.1\,kHz.}
\label{tab:stereo_reconstruction_44}
\resizebox{\textwidth}{!}{%
\begin{tabular}{llrrrrrrrr}
\toprule
Dataset & Model
& MEL$\downarrow$
& STFT$\downarrow$
& MFCC$\downarrow$
& Wave.$\downarrow$
& SI-SDR$\uparrow$
& SNR$\uparrow$
& PESQ$\uparrow$
& CCPC$\uparrow$ \\
\midrule

\multirow{2}{*}{Sounds}
& SAME-L
& 0.96 & 2.36 & 0.22 & \textbf{0.0449}
& 2.66 & \textbf{4.80} & -- & 0.8436 \\
& Ours
& \textbf{0.68} & \textbf{2.31} & \textbf{0.16} & 0.0475
& \textbf{2.73} & 4.75 & -- & \textbf{0.8575} \\
\cmidrule(lr){1-10}

\multirow{2}{*}{Music}
& SAME-L
& 0.88 & \textbf{2.64} & \textbf{0.24} & 0.0380
& 9.72 & 10.19 & -- & 0.9563 \\
& Ours
& \textbf{0.71} & 2.80 & 0.25 & \textbf{0.0313}
& \textbf{11.57} & \textbf{11.81} & -- & \textbf{0.9765} \\
\cmidrule(lr){1-10}

\multirow{2}{*}{Speech}
& SAME-L
& 0.81 & 1.85 & 0.17 & 0.0076
& 9.49 & 9.97 & 2.43 & 0.9981 \\
& Ours
& \textbf{0.34} & \textbf{1.23} & \textbf{0.06} & \textbf{0.0054}
& \textbf{11.78} & \textbf{12.07} & \textbf{4.47} & \textbf{0.9987} \\
\bottomrule
\end{tabular}
}
\end{table}

\begin{table}[H]
\centering
\caption{Stereo reconstruction results at 48\,kHz.}
\label{tab:stereo_reconstruction_48}
\resizebox{\textwidth}{!}{%
\begin{tabular}{llrrrrrrrr}
\toprule
Dataset & Model
& MEL$\downarrow$
& STFT$\downarrow$
& MFCC$\downarrow$
& Wave.$\downarrow$
& SI-SDR$\uparrow$
& SNR$\uparrow$
& PESQ$\uparrow$
& CCPC$\uparrow$ \\
\midrule

\multirow{2}{*}{Sounds}
& LTX 2.5
& 1.40 & 4.37 & 0.39 & 0.0888
& -48.30 & -2.42 & -- & 0.7311 \\
& Ours
& \textbf{0.68} & \textbf{2.33} & \textbf{0.16} & \textbf{0.0475}
& \textbf{2.74} & \textbf{4.75} & -- & \textbf{0.8576} \\
\cmidrule(lr){1-10}

\multirow{2}{*}{Music}
& LTX 2.5
& 1.32 & 4.05 & 0.36 & 0.1700
& -46.24 & -2.85 & -- & 0.7771 \\
& Ours
& \textbf{0.69} & \textbf{2.67} & \textbf{0.23} & \textbf{0.0311}
& \textbf{11.65} & \textbf{11.86} & -- & \textbf{0.9765} \\
\cmidrule(lr){1-10}

\multirow{2}{*}{Speech}
& LTX 2.5
& 1.18 & 3.39 & 0.30 & 0.0318
& -36.51 & -2.94 & 3.00 & 0.9881 \\
& Ours
& \textbf{0.36} & \textbf{1.42} & \textbf{0.07} & \textbf{0.0054}
& \textbf{11.78} & \textbf{12.07} & \textbf{4.47} & \textbf{0.9987} \\
\bottomrule
\end{tabular}
}
\end{table}

%% file: content/audio/generation.tex
\subsection{Generation results}
For generation experiments we use an open-source 0.6B diffusion transformer ~\citep{kandinsky5} adapted for audio. For every VAE we train the generator from scratch on an internal general-audio dataset under the same protocol for 60,000 steps on the latents of that VAE, and nothing else changes between comparisons.

To estimate human preferences we conduct blind, randomized pairwise listening tests on a fixed pool of 148 prompts spanning speech, music, and sounds. Listeners judge prompt following and audio quality separately and may mark both outputs as good or both as bad. Win rates give half credit for “both good” and exclude “both bad”.
Alongside listening tests, six objective generation metrics are reported: LAION CLAP for text–audio similarity, FAD (VGG) and FD (OPEN-L3) for distribution distances, the Content Enjoyment (CE) and Production Quality (PQ) scores of Audiobox Aesthetics, and, for speech prompts, WER and UTMOS.

Figure ~\ref{fig:audio-preferences} and Table ~\ref{tab:objective_generation_metrics} report the listening tests and the objective generation metrics.
Comparison with the open VAEs broken down by category, since this is where the domain coverage of the models differs most.
Our model is preferred over both VAEs in every category on both criteria. Against SAME-L it wins 55/68/64\% of the prompt-following and 71/98/93\% of the audio-quality comparisons for sounds, speech, and music.
Against the LTX 2.5 audio VAE it wins 74/60/67\% and 82/78/94\%.
On audio quality it thus wins more than 70\% of the comparisons against both VAEs in every category.
The objective metrics agree on AudioCaps and Seed-TTS Eval, where our model is best on nearly every metric, while on SongDescriber SAME-L and LTX 2.5 score higher on CLAP and FAD (VGG).

\begin{table}[H]
\centering
\begin{minipage}[t]{0.43\textwidth}
\vspace{0pt}
\centering
\includegraphics[width=\linewidth]{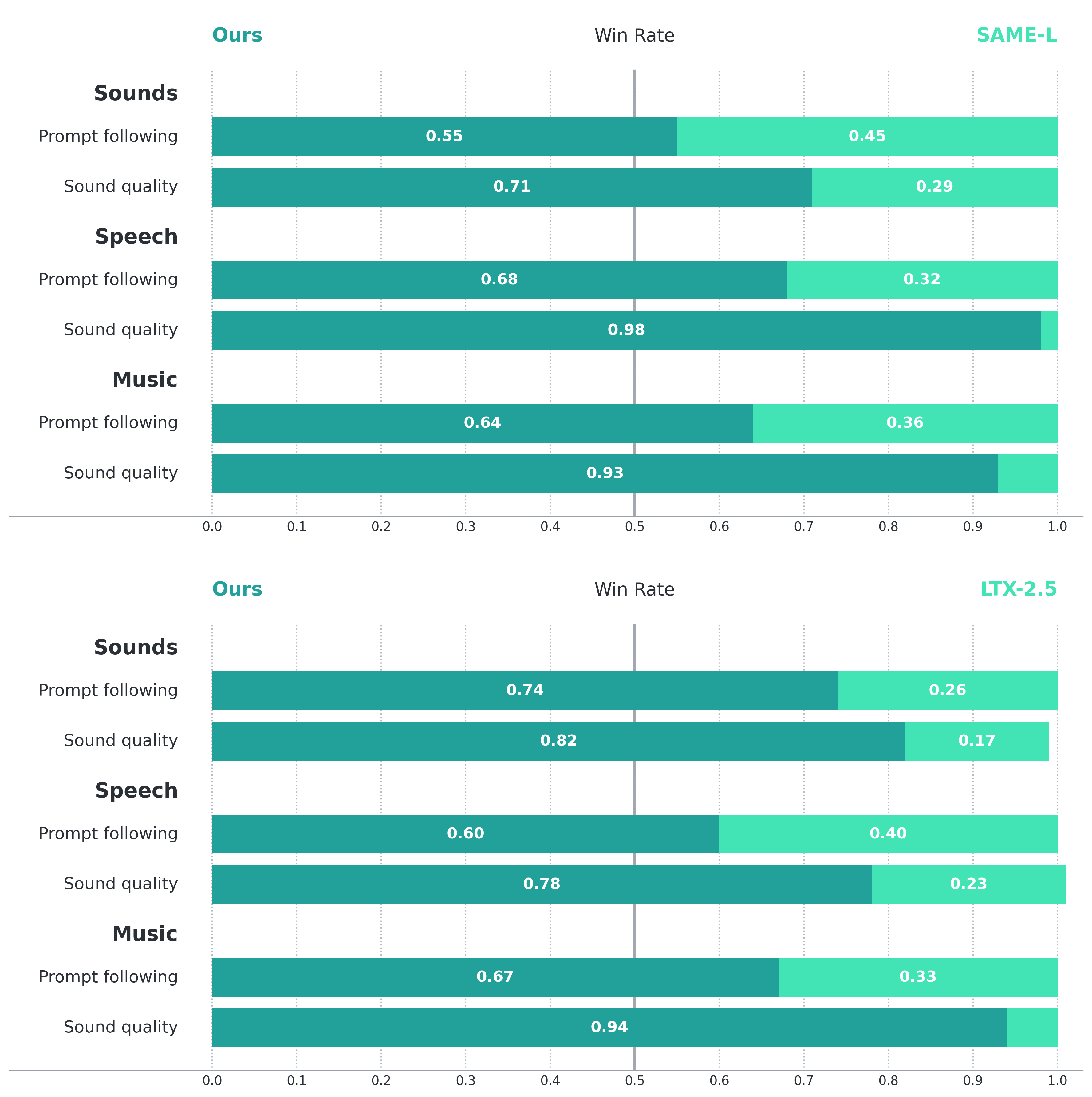}
\captionof{figure}{Pairwise audio preference win rates by category against SAME-L and LTX-2.5.}
\label{fig:audio-preferences}
\end{minipage}\hfill
\begin{minipage}[t]{0.55\textwidth}
\vspace{0pt}
\centering
\scriptsize
\begin{tabular}{lrrr}
\toprule
Metric & Ours & SAME-L & LTX 2.5 \\
\midrule

\multicolumn{4}{l}{\textit{AudioCaps test}} \\
LAION CLAP$\uparrow$   & \textbf{0.29}   & 0.25   & \underline{0.27} \\
CE$\uparrow$           & \textbf{4.33}   & 3.60   & \underline{4.04} \\
PQ$\uparrow$           & \textbf{6.52}   & 5.84   & \underline{6.36} \\
FD (OPEN-L3)$\downarrow$ & \textbf{218.34} & 232.87 & \underline{227.02} \\
FAD (VGG)$\downarrow$  & \textbf{4.27}   & 6.02   & \underline{5.13} \\
\midrule

\multicolumn{4}{l}{\textit{Song Describer}} \\
LAION CLAP$\uparrow$   & 0.26            & \textbf{0.30} & \underline{0.29} \\
CE$\uparrow$           & \textbf{7.57}   & 7.41   & \underline{7.56} \\
PQ$\uparrow$           & \textbf{8.14}   & 7.77   & \underline{8.10} \\
FD (OPEN-L3)$\downarrow$ & \textbf{114.36} & 127.87 & \underline{117.61} \\
FAD (VGG)$\downarrow$  & 2.20            & \underline{2.11} & \textbf{2.04} \\
\midrule

\multicolumn{4}{l}{\textit{Seed-TTS Eval}} \\
LAION CLAP$\uparrow$   & \textbf{0.17}   & \textbf{0.17} & 0.15 \\
CE$\uparrow$           & \textbf{5.93}   & 4.20   & \underline{5.42} \\
PQ$\uparrow$           & \textbf{7.25}   & 4.80   & \underline{6.32} \\
FD (OPEN-L3)$\downarrow$ & \underline{123.29} & 145.90 & \textbf{119.39} \\
FAD (VGG)$\downarrow$  & \textbf{4.51}   & 7.19   & \underline{5.80} \\
WER (\%)$\downarrow$   & \textbf{25.93}  & 58.97  & \underline{29.62} \\
\bottomrule
\end{tabular}
\captionof{table}{Objective generation metrics of the generators
trained on the three latents. \textbf{Bold}: best, \underline{underline}: second best.}
\label{tab:objective_generation_metrics}
\end{minipage}
\end{table}